\documentclass[11pt]{article}

\usepackage[preprint]{acl}

\usepackage{times}
\usepackage{latexsym}
\usepackage{amsmath}
\usepackage{amssymb}
\usepackage{booktabs}
\usepackage{pifont}
\usepackage{arydshln}

\usepackage[T1]{fontenc}

\usepackage[utf8]{inputenc}

\usepackage{microtype}

\usepackage{inconsolata}

\usepackage{graphicx}

\usepackage{listings}
\usepackage[most]{tcolorbox}
\usepackage{soul}
\usepackage{multirow}

\lstdefinestyle{promptStyle}{
  basicstyle=\fontsize{8}{9}\fontfamily{pcr}\selectfont,
  showstringspaces=false,
  breaklines=true,
  breakatwhitespace=false,
  breakindent=0pt,
  keepspaces=false,
  showspaces=false,
  escapeinside={(*@}{@*)},
  frame=none,
  aboveskip=0pt,
  belowskip=0pt,
}

\title{Practical Online KV Cache Compaction for LLM Agents:\\An Empirical Study}

\newcommand{\aspace}{\hspace{0.9em}}

\author{
Yujian Liu${^{1}}$\thanks{Equal contribution. Work done when Yujian, Jiabao, and Li were interning at LinkedIn.} \aspace Jiabao Ji${^{1*}}$ \aspace Li An${^{1*}}$ \aspace Rohit Jain${^{2}}$ \\ {\bf Gungor Polatkan${^{2}}$ \aspace Siyu Zhu${^{2}}$ \aspace Shiyu Chang${^{1}}$} \\
$^1$UC Santa Barbara \aspace $^{2}$LinkedIn \\
\texttt{\{yujianliu,jiabaoji,li\_an,chang87\}@ucsb.edu}}

\begin{document}
\maketitle
\begin{abstract}
LLM agents accumulate long trajectories of reasoning steps, tool calls, and environment feedback, making the KV cache a major inference bottleneck. KV cache compaction can reduce this cost, but most prior methods assume a static context where future queries are known or can be approximated offline. Agents instead require online compaction: new information must be compressed before future relevance is known, using proxy queries cheap enough for the inference path.
We study online compaction across token eviction (TE) and attention matching (AM), adapting both to compact agent turns and comparing cheap proxy sources such as boundary, repeat-prefill, and delayed future-generation queries. Experiments on \textsc{BrowseComp-Plus} and \textsc{WideSearch} show that immediate compaction often hurts performance, whereas delaying compaction to use the agent's future queries recovers much of the gap. Moreover, TE is often more robust than AM under imperfect proxies. Across models at different scales, TE preserves most of the accuracy while reducing KV cache by 80\%, and can improve throughput over the no compaction baseline. These results position proxy-query selection as a core design choice for practical online KV compaction.
\end{abstract}

\section{Introduction}
\label{sec:intro}

LLM agents are increasingly used for long-horizon tasks such as software engineering, deep research, web browsing, and personal assistance~\citep{anthropic2026opus47,openai2026gpt55,google2026gemini35,openclaw_github}. Their strength comes from maintaining an informative context: the model reasons, calls tools, observes results, and decides what to do next. This interaction pattern has enabled substantially more capable systems than single-shot prompting, but it also creates a direct inference bottleneck. Every generated reasoning step and every tool response is appended to the running context, and the model must retain a key-value (KV) cache for that growing history. As agent trajectories stretch across many turns, the cache can dominate memory use and increase the cost of each decoding step.

KV cache compaction is a natural way to reduce this cost. Instead of storing the full cache, a compaction method replaces a long sequence of cached keys and values with a shorter representation that approximately preserves the attention behavior of the original cache~\citep{liu2023scissorhandsexploitingpersistenceimportance,zhang2023h2oheavyhitteroracleefficient}. Much of the work compressing long prompts before answering studies a \emph{static} setting: a complete context is available before generation, the cache is compacted offline, and the resulting compact cache is then consumed by later queries. In such settings, signals about how the context will be used are often available or can be sampled offline. For example, query-aware token eviction methods can use an observation window containing the question or continuation that will read the context~\citep{li2024snapkvllmknowslooking,cai2025pyramidkvdynamickvcache}. Other approaches can further spend additional computation to create training signals before inference, either by generating proxy queries for fitting a compact cache or by generating synthetic conversations and distilling them into a trainable cache~\citep{zweiger2026fastkvcompactionattention,eyuboglu2025cartridgeslightweightgeneralpurposelong}.

Agent trajectories, however, differ from this static setting. An agent does not begin with a fixed document to compress; it gradually constructs its context through interactions with an environment. Search results, file diffs, and execution traces arrive throughout the trajectory and are incorporated into the live context. Compaction in this regime is therefore inherently \emph{online}: newly collected information must be compacted before the full future trajectory is known. Two properties distinguish it from static compaction. \textbf{First, future relevance is not yet observable.} When a turn is compacted, the information needed from it several turns later may differ from what that turn or even the immediately following turn would emphasize. A compaction signal derived only from the current context may therefore poorly represent how the compacted cache will be used later. \textbf{Second, the compaction procedure must fit within a reasonable time budget.} Because compaction is performed inside the agent loop, both obtaining a signal about what information to preserve and executing the compaction algorithm contribute directly to the agent's runtime. Procedures that rely on additional rollouts, long synthetic continuations, or expensive optimization may be feasible offline but can add substantial latency inside a live agent trajectory. Together, these properties create an underexplored design space for adapting existing compaction methods: which available signals should guide compaction, and how signal quality and runtime cost interact with the compaction method.

To study this design space, we present a systematic empirical evaluation of how existing KV cache compaction methods behave when adapted to agent tasks. We focus on two widely used sequence-level compaction families, both of which depend on query vectors that represent how the compacted cache will later be read. We refer to these as \emph{proxy queries}. Token eviction (TE) uses proxy queries to score cached positions by their induced attention mass, then keeps the original KVs at the highest-scoring positions~\citep{li2024snapkvllmknowslooking}. Attention matching (AM) uses the same proxy-query-based key selection, but additionally fits an additive attention bias and reconstructed values so that attention outputs against the compacted cache better match those against the full cache~\citep{zweiger2026fastkvcompactionattention}. We adapt both families to the online agent setting by compacting completed turns as the trajectory unfolds, where proxy queries are derived from signals available at or near the completed turns. Our main analysis studies the interaction between compaction family and proxy source, including current-turn boundary queries, lightweight repeat-prefill queries, and deferred future-turn queries obtained from the agent's own subsequent generation. The resulting experiments map which proxy sources are useful, when deferring compaction helps, and when adding more proxy sources hurts.

We evaluate on \textsc{BrowseComp-Plus}~\citep{chen2025browsecompplusfairtransparentevaluation} and \textsc{WideSearch}~\citep{wong2025widesearchbenchmarkingagenticbroad}, using \texttt{Qwen3.5} and \texttt{Gemma-4} models at multiple scales. Experiments show that immediate compaction often degrades task performance, whereas delaying compaction to use queries from the agent's own subsequent generation consistently recovers much of the gap. Despite its much simpler selection-only design, TE is surprisingly robust to imperfect proxies and remains competitive with AM across proxy, delay, and compaction-budget choices. On \texttt{Qwen3.5-27B} and \texttt{Gemma-4-31B}, delayed compaction at a ratio of 0.2 preserves most of the no-compaction accuracy while reducing peak KV footprint by up to 3.5$\times$ and 2.7$\times$, and increasing serving throughput by up to 4.2$\times$ and 1.7$\times$, respectively. Together, these results identify proxy-query selection as core design choices for practical online KV compaction.
\section{Related Work}
\label{sec:related_work}

\paragraph{Token-level KV eviction and query-aware sparsity.}
A large body of work reduces KV memory by retaining only selected cached tokens. Attention-based eviction methods exploit concentration, persistence, or structural attention patterns to keep heavy-hitter, sink, recent, or head-specific tokens~\citep{liu2023scissorhandsexploitingpersistenceimportance,zhang2023h2oheavyhitteroracleefficient,xiao2024efficientstreaminglanguagemodels,ge2024modeltellsdiscardadaptive}. Query- or layer-aware methods further score historical tokens from an observation window or allocate budgets across layers~\citep{li2024snapkvllmknowslooking,tang2024questqueryawaresparsityefficient,yang2024pyramidinferpyramidkvcache,cai2025pyramidkvdynamickvcache}.
Expected attention similarly estimates importance from a future-query distribution when future attention is unavailable~\citep{devoto2025expectedattentionkvcache}. These methods show that query-conditioned signals are useful for KV reduction. In agent trajectories, however, future queries are endogenous to the model's own later actions. A compaction decision made after one tool interaction must therefore preserve information for future states of the trajectory that do not yet exist.

\paragraph{Learned and optimized compact representations.}
Other work replaces token selection with learned or optimized compact states. Prompt- and context-compression methods summarize contexts into soft tokens, recurrent states, or compressed activation slots~\citep{mu2024learningcompresspromptsgist,chevalier2023adaptinglanguagemodelscompress,ge2024incontextautoencodercontextcompression,zhang2024longcontextcompressionactivation,chu2026kwaisummaryattentiontechnical}.
Cartridges trains compact KV representations through offline self-study, requiring synthetic conversations and gradient-based optimization~\citep{eyuboglu2025cartridgeslightweightgeneralpurposelong}. Attention matching fits a compact cache to match full-cache attention outputs on proxy queries~\citep{zweiger2026fastkvcompactionattention}. We use attention matching as one compaction family in our study. These methods are usually static: the context, task distribution, or proxy queries can be prepared before deployment. We instead study compaction when proxy queries must be obtained cheaply during an ongoing agent trajectory.

\paragraph{Multi-turn and agent KV management.}
Several recent works study KV reuse or compression in multi-turn settings. KVzip motivates query-agnostic compression for caches that may be reused by many future queries, and \textsc{SCBench} shows that sublinear-memory methods can degrade under multi-turn KV-cache lifecycles~\citep{kim2025kvzipqueryagnostickvcache,li2025scbenchkvcachecentricanalysis}. Serving systems improve multi-turn or agent efficiency by reusing, scheduling, or retaining KV states across conversation and tool-execution boundaries~\citep{gao2024costefficientlargelanguagemodel,li2026continuumefficientrobustmultiturn}. These works are complementary to ours. They treat the growing interaction history as a systems object to cache, reuse, or schedule, whereas we ask which information inside each newly completed agent turn should be preserved when that turn is compacted and then frozen. \textsc{SCBench}, for example, studies shared-context settings in which multiple requests query a long context that is already available. In our setting, the context is produced online: search results, file contents, and other observations are appended only after the agent issues the corresponding actions. This makes online compaction a decision about an evolving trajectory, not only reuse of a fixed context.

\paragraph{Orthogonal KV-compression dimensions.}
KV memory can also be reduced along axes orthogonal to sequence length: quantization lowers key/value precision~\citep{liu2023kivi,hooper2025kvquant10millioncontext,kang2024gearefficientkvcache}; low-rank methods compress or offload KV states through lower-dimensional representations~\citep{zhang2024lorclowrankcompressionllms,sun2025shadowkvkvcacheshadows,xu2025thinkthinnerkeycache}; and architectural methods use grouped or latent KV representations~\citep{ainslie2023gqatraininggeneralizedmultiquery,deepseekai2024deepseekv2strongeconomicalefficient}. These approaches address a different axis of the memory problem and can be combined with compaction on sequence length.
\section{Design Space of Online KV Compaction}
\label{sec:prelim}

We first describe the online KV compaction problem. The goal is not to propose a new algorithm or a single final recipe, but to isolate which choices matter when a compactor is placed inside an agent loop. We begin by formalizing the two compaction families studied throughout the paper, then describe how we adapt them from a static prompt setting to online agent trajectories.

\subsection{Static KV Compaction}
\label{sec:prelim_static}

\begin{figure*}[t]
  \centering
  \includegraphics[width=0.95\linewidth]{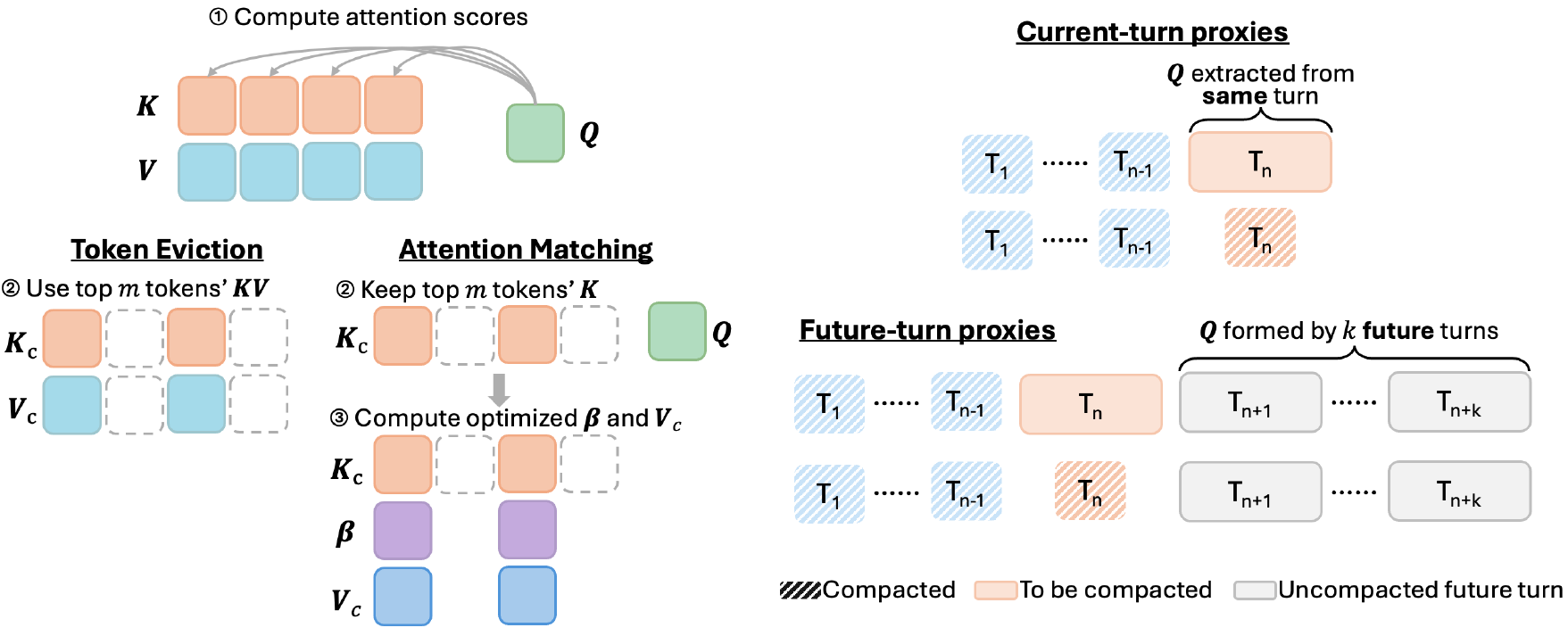}
  \caption{Overview of the online KV compaction design space. \textbf{Left}: TE and AM both score cached tokens using proxy queries. TE stores the selected original keys and values, while AM keeps the selected keys and optimizes the attention bias and compacted values. \textbf{Right}: current-turn proxies are available when the turn finishes, whereas future-turn proxies delay compaction until $k$ later turns provide additional queries.}
  \label{fig:online_compaction_design}
\end{figure*}

Consider one attention layer and one KV head. We omit the usual $1/\sqrt{d}$ attention scale for notational simplicity. Let $\mathbf{K},\mathbf{V} \in \mathbb{R}^{n \times d}$ denote the full key and value cache for a context of length $n$, and let $\mathbf{Q} \in \mathbb{R}^{q \times d}$ denote a set of $q$ proxy queries. The full-cache attention output on these queries is
\begin{equation}
  \mathbf{A}(\mathbf{Q};\mathbf{K},\mathbf{V})
  =
  \mathrm{softmax}\!\left(\mathbf{Q}\mathbf{K}^\top\right)
  \mathbf{V}.
\end{equation}
A compactor constructs a shorter cache of length $m \ll n$ whose attention outputs approximate $\mathbf{A}(\mathbf{Q};\mathbf{K},\mathbf{V})$. We focus on TE and AM because they are widely used and, crucially for the online setting, their compaction steps can be executed efficiently once the proxy queries are available. The left side of Figure~\ref{fig:online_compaction_design} illustrates these two families.

\paragraph{Token eviction.}
Token eviction (TE)~\citep{li2024snapkvllmknowslooking} keeps original KV entries at selected positions. The selection is driven by attention scores induced by the proxy queries. Let
\begin{equation}
  \alpha_{ij}
  =
  \left[
  \mathrm{softmax}\!\left(\mathbf{Q}\mathbf{K}^{\top}\right)
  \right]_{ij}
\end{equation}
be the attention weight from proxy query $i$ to cached position $j$. We score each cached position by its root-mean-square attention mass,
\begin{equation}
  s_j(\mathbf{Q},\mathbf{K})
  =
  \left(\frac{1}{q}\sum_{i=1}^{q}\alpha_{ij}^2\right)^{1/2}.
  \label{eq:te_score}
\end{equation}
TE selects the highest-scoring positions and stores the corresponding original keys and values:
\begin{equation}
  S = \mathrm{Top}\text{-}m\{s_j\}_{j=1}^{n},
  \quad
  \mathbf{K}_C = \mathbf{K}_S,\quad
  \mathbf{V}_C = \mathbf{V}_S,
  \label{eq:te_cache}
\end{equation}
where $\mathbf{K}_C,\mathbf{V}_C \in \mathbb{R}^{m \times d}$. TE is thus a pure selection method: after selecting $S$, the stored keys and values are unchanged.

\paragraph{Attention matching.}
Attention matching (AM)~\citep{zweiger2026fastkvcompactionattention} uses the same proxy-query-based key selection as TE, setting $\mathbf{K}_C = \mathbf{K}_S$. It then fits an additive bias $\boldsymbol{\beta} \in \mathbb{R}^{m}$ and compact values $\mathbf{V}_C \in \mathbb{R}^{m \times d}$. It consists of two stages. First, $\boldsymbol{\beta}$ is
chosen so that the selected keys account for the full cache's unnormalized attention mass. Let $\mathbf{1}_m \in \mathbb{R}^{m}$ and $\mathbf{1}_n \in \mathbb{R}^{n}$ be all-ones vectors:
\begin{equation}
  \min_{\boldsymbol{\beta}}
  \left\|
  \exp(\mathbf{Q}\mathbf{K}_C^\top+\boldsymbol{\beta})\mathbf{1}_m
  -
  \exp(\mathbf{Q}\mathbf{K}^{\top})\mathbf{1}_n
  \right\|_2^2,
  \label{eq:am_beta}
\end{equation}
where $\boldsymbol{\beta}$ is broadcast across queries. Then, holding $\boldsymbol{\beta}$ fixed, AM fits values to match the full-cache attention outputs. Define
\begin{equation}
  \widehat{\mathbf{A}}(
  \mathbf{Q};\mathbf{K}_C,\boldsymbol{\beta},\mathbf{V}_C)
  \triangleq
  \mathrm{softmax}\!\left(
  \mathbf{Q}\mathbf{K}_C^\top+\boldsymbol{\beta}
  \right)\mathbf{V}_C ,
\end{equation}
AM solves
\begin{equation}
  \min_{\mathbf{V}_C}
  \left\|
  \widehat{\mathbf{A}}(
  \mathbf{Q};\mathbf{K}_C,\boldsymbol{\beta},\mathbf{V}_C)
  -
  \mathbf{A}(\mathbf{Q};\mathbf{K},\mathbf{V})
  \right\|_F^2 .
  \label{eq:am_objective}
\end{equation}
The bias $\boldsymbol{\beta}$ changes how much attention each selected key receives, while $\mathbf{V}_C$ is allowed to differ from the original selected values. Thus TE commits to the original KV, whereas AM keeps the selected keys but fits the attention bias and values to match full-cache outputs on the proxy queries.

This formulation is applied independently to each layer and KV head. In our experiments, both \texttt{Qwen3.5}~\citep{qwenteam2026qwen35omnitechnicalreport} and \texttt{Gemma-4}~\citep{gemma4_model_card_2026} are hybrid-attention models, so we compact only their full-attention layers and leave non-full-attention states unchanged; implementation details are given in Appendix~\ref{sec:appendix}.

\subsection{Online Compaction for Agent Trajectories}
\label{sec:prelim_online}

In an agent trajectory, the context is not fixed before generation. The model starts from a prefix containing the system prompt, tool definitions, and user query. It then repeatedly generates assistant messages, emits tool calls, receives tool responses, and continues from the expanded context. We segment this history into turns:
\[
  P,\ T_1,\ T_2,\ldots,T_t,\ldots
\]
where $P$ is the uncompacted prefix and each $T_t$ contains the assistant-side generation for that step and the resulting tool response. Online compaction compresses completed turns as they arrive. Once turn $T_t$ is compacted, future generations attend to the compacted representation and discard the original KV; the compacted turn is then frozen and is not re-optimized or re-compacted later.

This setup changes the role of proxy queries. In the static setting, proxy queries can be drawn from known questions or offline synthetic data. In the online setting, the future queries that will read $T_t$ are not yet available when $T_t$ finishes. We therefore study two cheap proxy families that do not require extra rollouts, as summarized in Figure~\ref{fig:online_compaction_design} right.

\paragraph{Current-turn proxies.}
These proxies are available immediately when $T_t$ completes, so they allow immediate compaction. We consider two ways to obtain such proxies.
\ding{182} \textbf{Boundary queries} use the model's query vectors at structural closing or transition tokens from the actual trajectory. For example, for \texttt{Qwen3.5}, we extract query vectors from the closing token \texttt{<|im\_end|>}. These queries are cheap and require no extra forward pass, but they may not reflect what future turns will need from $T_t$.
\ding{183} \textbf{Repeat-prefill queries} are inspired by query-agnostic KV compaction in KVzip~\citep{kim2025kvzipqueryagnostickvcache}. We append a reconstruction prompt after the previous turn's content and teacher-force the model to repeat the completed turn:
\begin{quote}
\small
\texttt{<previous context>}\\
\texttt{Let me repeat the previous thinking, tool call, and tool response.}\\
\texttt{<repeated context>}
\end{quote}
We extract proxy queries only from the teacher-forced repeated content. The repeated thinking and tool-call text provides proxies for compacting the assistant generation, while the repeated tool response provides proxies for compacting the tool-response segment.

\paragraph{Future-turn proxies.}
The second proxy family delays compaction in order to use real future queries. To compact $T_t$ with a delay of $k$ turns, the system keeps $T_t$ in raw form while generating turns $T_{t+1},\ldots,T_{t+k}$. During these generations, the model attends to the uncompressed prefix $P$, earlier frozen compacted turns, the raw cache for $T_t$, and any newer turns that are still inside the delay window. By default, we record the query vectors produced during the assistant generations in $T_{t+1},\ldots,T_{t+k}$ and use them as proxies when compacting $T_t$ after $T_{t+k}$ finishes. We also ablate adding the query vectors from the corresponding tool-response prefill tokens as an additional proxy source. From turn $T_{t+k+1}$ onward, $T_t$ is read only through its frozen compacted representation. Thus a larger delay provides proxy queries that are closer to the way later computation actually reads $T_t$, but it also postpones the memory and compute savings because more raw turns must remain in the cache.
\section{What Matters for Online KV Compaction}
\label{sec:online_results}

\subsection{Setup}
\label{sec:online_results_setup}

We evaluate on two complementary agentic-search benchmarks. \textsc{BrowseComp-Plus}~\citep{chen2025browsecompplusfairtransparentevaluation} is a fixed-corpus benchmark for deep-search agents: each example asks a difficult information-seeking question with a verifiable final answer, and solving it requires iteratively retrieving evidence from the corpus rather than relying on parametric knowledge. \textsc{WideSearch}~\citep{wong2025widesearchbenchmarkingagenticbroad} instead targets broad information seeking. It requires the agent to collect many atomic facts and organize them into a well-structured table, so completeness matters as much as correctness.

In both settings the agent starts from the system prompt and user question, then repeatedly generates tool calls, observes tool outputs, and either continues searching or returns its final answer. For \textsc{BrowseComp-Plus}, the agent queries a local FAISS search server with \texttt{Qwen3-Embedding-4B}~\citep{zhang2025qwen3embeddingadvancingtext} retrieval. For \textsc{WideSearch}, the agent searches the open internet through Bing Web Search API. We study two model families: \texttt{Qwen3.5} and \texttt{Gemma-4}, both of which use hybrid attention. In this section, we report performance of the 4B models and defer results of larger models to Section~\ref{sec:additional_analysis}. We compact assistant generation and tool response separately, while preserving structural boundary tokens needed by the chat template. We score both benchmarks with a \texttt{Qwen3.5-397B} judge. For \textsc{BrowseComp-Plus} we report answer accuracy under the official grader template; for \textsc{WideSearch} we report the item-level F1 for the final table, which credits partially collected atomic facts.

\subsection{Proxy-Query Ablation}
\label{sec:online_results_proxy}

We first fix the compaction ratio to 0.2, reducing each compactable segment to 20\% of its original length, and compare how different proxy sources affect performance. This section focuses on task performance; we study runtime savings in Section~\ref{sec:additional_efficiency}. Table~\ref{tab:proxy_ablation} organizes the results into current-turn proxies and one-turn-delayed future proxies. Corresponding 95\% bootstrap confidence intervals are reported in Appendix Table~\ref{tab:appendix_proxy_ci}.

\begin{table*}[t]
\centering
\resizebox{0.98\textwidth}{!}{%
\begin{tabular}{ll c cc cc cc cc}
\toprule
 & \multirow{2}{*}{Proxy source} & \multirow{2}{*}{Delay} &
\multicolumn{4}{c}{\textsc{BrowseComp-Plus}} &
\multicolumn{4}{c}{\textsc{WideSearch}} \\
\cmidrule(lr){4-7} \cmidrule(lr){8-11}
 & & &
\multicolumn{2}{c}{\texttt{Qwen3.5-4B}} &
\multicolumn{2}{c}{\texttt{Gemma-4-E4B}} &
\multicolumn{2}{c}{\texttt{Qwen3.5-4B}} &
\multicolumn{2}{c}{\texttt{Gemma-4-E4B}} \\
\cmidrule(lr){4-5} \cmidrule(lr){6-7} \cmidrule(lr){8-9} \cmidrule(lr){10-11}
 & & & Acc. & Turns & Acc. & Turns & F1 & Turns & F1 & Turns \\
\midrule
 & No compaction & -- & 46.00 & 20 & 33.00 & 13 & 44.55 & 24 & 31.94 & 9 \\
\midrule
\multicolumn{11}{c}{\textbf{Current-turn proxies}} \\
\midrule
AM & Repeat-prefill & 0 & 31.25 & 31 & 11.25 & 9 & \textbf{30.95} & 28 & 13.63 & 8 \\
TE & Repeat-prefill & 0 & 32.75 & 51 & 9.00 & 12 & 24.56 & 82 & \textbf{17.19} & 11 \\
TE & Boundary & 0 & \textbf{45.25} & 40 & \textbf{21.50} & 12 & 24.29 & 50 & 15.73 & 11 \\
\midrule
\multicolumn{11}{c}{\textbf{One-turn-delayed future proxies}} \\
\midrule
AM & Assistant generation & 1 & 39.75 & 33 & \textbf{27.50} & 10 & \textbf{39.59} & 33 & 20.17 & 9 \\
AM & + repeat-prefill & 1 & \textbf{43.25} & 30 & 24.25 & 10 & 38.83 & 29 & \textbf{21.49} & 9 \\
AM & + repeat-prefill + tool response & 1 & 43.25 & 30 & 27.50 & 9 & 37.39 & 29 & 20.17 & 9 \\
\hdashline
TE & Assistant generation & 1 & \textbf{44.00} & 38 & \textbf{27.50} & 10 & 37.34 & 39 & 25.90 & 11 \\
TE & + boundary & 1 & 43.00 & 37 & 27.00 & 11 & 38.09 & 35 & \textbf{26.53} & 11 \\
TE & + boundary + tool response & 1 & 42.25 & 39 & 26.75 & 11 & \textbf{39.14} & 35 & 25.39 & 10 \\
\bottomrule
\end{tabular}
}
\caption{Proxy-source ablation at compaction ratio 0.2 on \textsc{BrowseComp-Plus} and \textsc{WideSearch}. For \textsc{BrowseComp-Plus}, ``Acc.'' is final answer accuracy; for \textsc{WideSearch}, ``F1'' is the mean item-level table F1. ``Turns'' is the median number of agent turns.}
\label{tab:proxy_ablation}
\end{table*}

\paragraph{Which current-turn proxy should we use?}
We compare repeat-prefill and boundary queries for TE, and repeat-prefill for AM. Boundary queries are not evaluated for AM because AM fits bias and value parameters against proxy targets; a single boundary query provides a poorly constrained optimization target. For TE, the stronger current-turn proxy depends on the benchmark. On \textsc{BrowseComp-Plus}, replacing repeat-prefill with boundary queries improves \texttt{Qwen3.5-4B} from 32.75\% to 45.25\% and \texttt{Gemma-4-E4B} from 9.00\% to 21.50\%. On \textsc{WideSearch}, however, the two proxies perform similarly. Boundary queries can therefore act as useful aggregation points, but their advantage is not universal. Their strong performance on \textsc{BrowseComp-Plus} is also consistent with recent work that uses attention to an end-of-thinking token to identify important reasoning tokens, as well as analyses showing that punctuation and other structural tokens can form semantically meaningful attention sinks~\citep{choi2025thinkclearlyimprovingreasoning,zhang2025attentionsinkscatchtag}.

\paragraph{Does delaying compaction help?}
Future-turn proxies use actual assistant-generation queries from the next turn, so they are closer to the way the compacted turn will be read later. Across all four model and benchmark pairs, using one-turn-delayed assistant-generation queries improves over immediate repeat-prefill for both AM and TE. For TE, delayed queries also outperform boundary queries in three of the four pairs. The only exception is \texttt{Qwen3.5-4B} on \textsc{BrowseComp-Plus}, where immediate boundary queries reach 45.25\% accuracy, slightly above the 44.00\% obtained with delayed queries. The consistency of the results shows that even a one-turn delay generally provides a more informative compaction signal than proxies constructed from the current turn alone.

\paragraph{Should we combine proxy sources?}
We next add the best available current-turn proxy to the future-turn proxy: repeat-prefill for AM and boundary queries for TE. We then add future tool-response queries as a third source. The combination rule differs by compaction family. For TE, each proxy source selects a fixed portion of the token budget. For example, with two sources, half the selected tokens come from future-turn queries and half from boundary queries. For AM, we downsample each additional proxy source to the same length as the future turn source, then concatenate all sources as the optimization target. The results show that more proxy sources do not automatically help. For TE, all three configurations are within 2 points of one another for every model and benchmark pair, suggesting TE's robustness to variations in proxy source, so long as the future assistant generation is available. Overall, the configurations perform comparably, and adding current-turn or tool-response queries provides no reliable gain.

\paragraph{How does compaction change agent behavior?}
Table~\ref{tab:proxy_ablation} also shows that compaction changes the agent's behavior, not only its final performance. Across both benchmarks, \texttt{Qwen3.5-4B} tends to lengthen its trajectories under compaction. Its no-compaction baselines use medians of 20 turns on \textsc{BrowseComp-Plus} and 24 turns on \textsc{WideSearch}, whereas the delayed configurations use 30-39 and 29-39 turns, respectively. In contrast, \texttt{Gemma-4-E4B} remains close to its no-compaction trajectory length on both benchmarks. The longer \texttt{Qwen3.5-4B} trajectories suggest that the model may compensate for weakened context by issuing additional searches and recovering missing evidence through the environment. We examine this hypothesis directly in Section~\ref{sec:additional_behavior}. This cross-model difference reinforces that online compaction should be evaluated as an agent-level intervention, not only as an attention-approximation problem.

\begin{figure*}[t]
  \centering
  \includegraphics[width=0.8\linewidth]{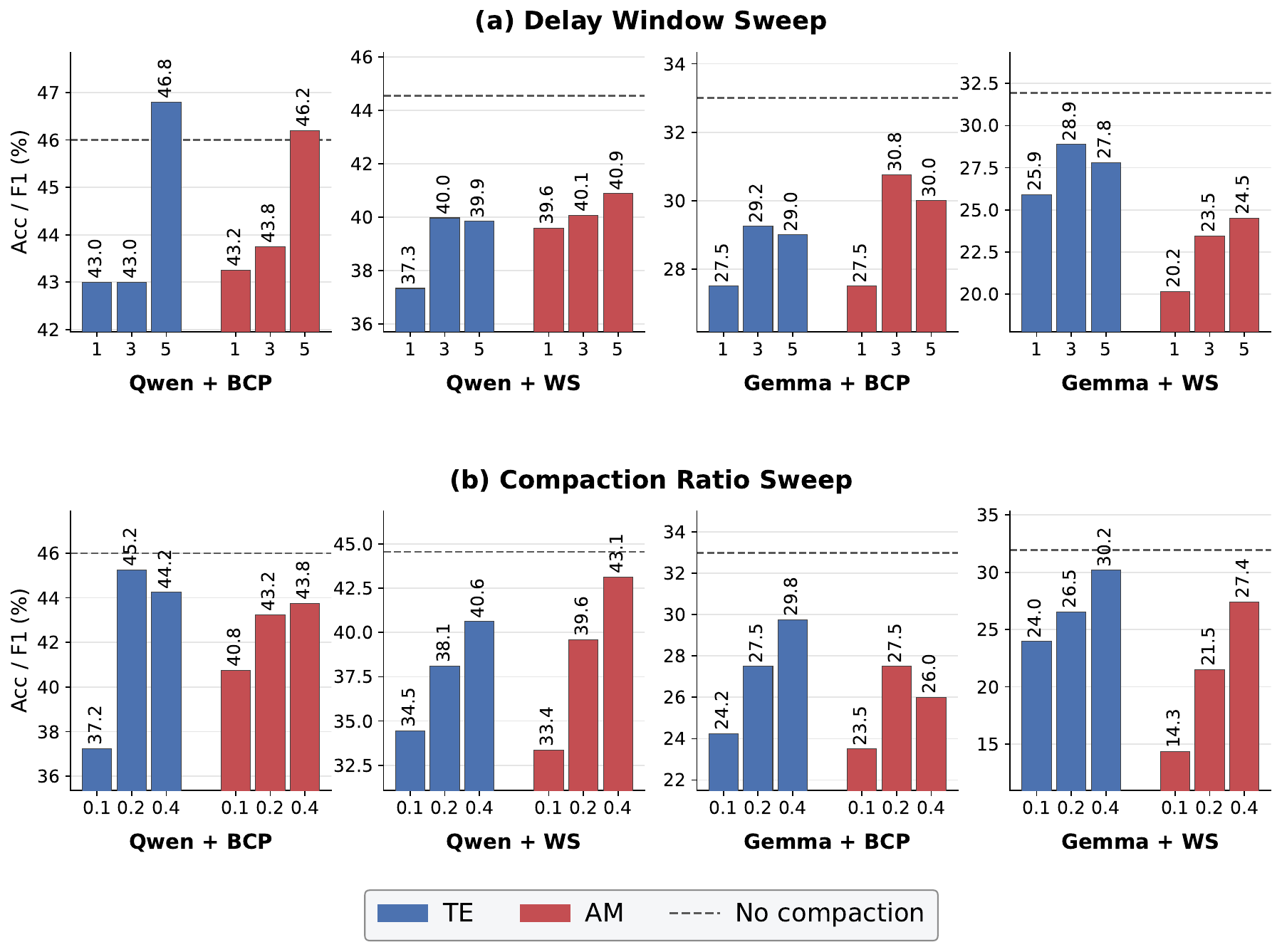}
  \caption{Sensitivity to compaction budget and delay window. The top row varies the delay window over $\{1,3,5\}$ turns at compaction ratio $0.2$; the bottom row varies the compaction ratio over $\{0.1,0.2,0.4\}$ using the stronger short-delay setting for each model and method pair. Each panel fixes one model (\texttt{Qwen3.5-4B} or \texttt{Gemma-4-E4B}) and benchmark (\textsc{BrowseComp-Plus} or \textsc{WideSearch}). Dashed lines mark the no-compaction baselines.}
  \label{fig:budget_delay_sweeps}
\end{figure*}

\subsection{Sensitivity to Compaction Budget and Delay}
\label{sec:online_results_sweep}

The previous ablation fixes the compaction ratio and uses at most a one-turn delay. We next characterize how performance varies with the memory budget and delay window. For each model--benchmark--method combination, we select the best-performing proxy strategy from Table~\ref{tab:proxy_ablation} and sweep either the compaction ratio or delay. Figure~\ref{fig:budget_delay_sweeps} shows the results. For the delay, we fix the compaction ratio to 0.2 and vary the delay over 1, 3, and 5 turns. For the compaction-ratio, we vary the ratio over 0.1, 0.2, and 0.4, using the stronger short-delay setting between immediate compaction and a one-turn delay for each combination.

The delay sweep shows that additional future context helps, although the gains are not always monotonic. For all combinations, longer delays (3 or 5 turns) outperform the one-turn delay. On \textsc{BrowseComp-Plus} with \texttt{Qwen}, both methods benefit most from a five-turn delay, slightly exceeding the no-compaction baseline. In the other settings, gains tend to peak or flatten earlier. Thus, observing more of the agent's future queries is generally useful, but the best delay depends on the model, benchmark, and compaction method.

The compaction-ratio sweep shows a clearer overall relationship with memory budget. Increasing the ratio from 0.1 to 0.2 improves every model--benchmark--method combination, confirming that a ratio of 0.1 is too aggressive in these settings. Increasing the ratio further to 0.4 improves six of the eight combinations. In particular, performance increases monotonically with budget on \textsc{WideSearch}. The exceptions occur on \textsc{BrowseComp-Plus}, where \texttt{Qwen3.5-4B} TE and \texttt{Gemma-4-E4B} AM both peak at ratio 0.2. Thus, a larger compact cache usually improves performance, particularly for broad search.

\begin{table*}[t]
\centering
\resizebox{0.95\textwidth}{!}{%
\begin{tabular}{ll c r r r r}
\toprule
Model & Method & Acc. & Avg. turns & Peak KV (k tokens) & Serving batch & Throughput (q/h) \\
\midrule
\multirow{3}{*}{\texttt{Qwen3.5-27B}}
 & No compaction & 52.50 & 18 & 612.9 & 8 & 217 \\
 & AM & 51.00 & 28 & 172.8 & 32 & \textbf{918} \\
 & TE & 52.00 & 31 & 175.2 & 32 & 717 \\
\midrule
\multirow{3}{*}{\texttt{Gemma-4-31B}}
 & No compaction & 48.75 & 13 & 272.1 & 8 & 217 \\
 & AM & 50.00 & 17 & 99.6 & 16 & \textbf{364} \\
 & TE & 50.75 & 18 & 121.3 & 16 & 324 \\
\bottomrule
\end{tabular}
}
\caption{Task performance and simulated serving efficiency on larger models. Accuracy, average turns, and trajectory lengths come from batch-size-1 runs on \textsc{BrowseComp-Plus}. Peak KV is the maximum per-turn KV length. Serving batch is the batch used for the throughput simulation. All compaction settings use one-turn-delayed assistant-generation queries and a compaction ratio of 0.2.}
\label{tab:larger_model_efficiency}
\end{table*}

\paragraph{Takeaways.}
Across these experiments, current-turn proxy choice is benchmark-dependent, and combining proxy sources provides no reliable gain. In contrast, delayed future-turn queries improve performance consistently, although the best delay varies across settings. Increasing the memory budget also usually helps, especially on \textsc{WideSearch}, but a larger compact cache is not uniformly better. Finally, AM's additional optimization is not automatically beneficial under online constraints. Despite being much simpler, TE remains surprisingly competitive across proxy, delay, and budget choices, suggesting that AM's richer optimization provides limited benefit when the available proxy queries imperfectly represent how the cache will later be used.
\section{Additional Analyses}
\label{sec:additional_analysis}

The study in Section~\ref{sec:online_results} focuses on answer accuracy and identifies strategies that preserve most of the no-compaction performance on smaller models. We next study two complementary questions. First, do these strategies preserve task performance and improve serving efficiency on larger models? Second, how does compaction change agent behavior beyond final answer accuracy?

\subsection{Task Performance and Serving Efficiency on Larger Models}
\label{sec:additional_efficiency}
We extend our evaluation to \texttt{Qwen3.5-27B} and \texttt{Gemma-4-31B}, substantially larger models from both architectural families. We evaluate on \textsc{BrowseComp-Plus} at a compaction ratio of 0.2. We compare no compaction with AM and TE using one-turn-delayed assistant-generation queries in this section. The complete proxy-source ablation results, including 95\% confidence intervals, are reported in Appendix Table~\ref{tab:appendix_larger_model_results}, which demonstrate a similar trend as the 4B models.

Table~\ref{tab:larger_model_efficiency} reports both task performance and serving efficiency. We generate agent trajectories at batch size 1 using our Hugging Face evaluation code and use these runs to measure answer accuracy and trajectory length. To estimate how the same workloads would execute on an optimized inference engine, we extract the KV length and number of generated tokens at every turn and replay these trajectory shapes using decode latencies measured with SGLang~\cite{zheng2024sglangefficientexecutionstructured}. This connects the task outcomes produced by our evaluation code to the memory and latency characteristics of a production-oriented serving stack. We use each method's peak KV footprint to determine the largest admissible batch and report the resulting decode throughput. Appendix~\ref{sec:appendix_serving_simulation} provides the complete simulation setup.

Table~\ref{tab:larger_model_efficiency} shows that both methods retain most of the no-compaction task performance at larger scale. On \texttt{Qwen3.5-27B}, no compaction obtains 52.50\% accuracy, compared with 51.00\% for AM and 52.00\% for TE. On \texttt{Gemma-4-31B}, both methods slightly exceed the baseline, which partly reflects the fact that some no-compaction trajectories run out of memory even at batch size 1, whereas compaction keeps their KV caches within the limit.

For \texttt{Qwen3.5-27B}, compaction reduces peak KV from 612.9K tokens to about 175K tokens. This raises the guaranteed serving batch from 8 to 32. As a result, aggregate throughput increases from 217 queries/hour without compaction to 918 with AM and 717 with TE, corresponding to 4.2$\times$ and 3.3$\times$ improvements. On \texttt{Gemma-4-31B}, peak KV falls from 272.1K tokens to 99.6K with AM and 121.3K with TE. The serving batch increases from 8 to 16, raising throughput by 1.7$\times$ and 1.5$\times$, respectively. The gains are smaller than for \texttt{Qwen3.5-27B}, but the same pattern holds across both models: compaction reduces the memory footprint enough to support more concurrent requests. Thus, even though the compacted agents take more turns on average, the increase in concurrency offsets the additional decode work, yielding higher overall throughput in queries per hour.

\begin{table}
\centering
\begin{tabular}{lcc}
\toprule
Method & NN Sim. & Duplicate \\
\midrule
No compaction & 0.72 & 11.7$\%$ \\
TE-boundary & 0.78 & 21.3$\%$ \\
AM-delay+RP & 0.75 & 16.7$\%$ \\
\bottomrule
\end{tabular}
\caption{Search-query repetition within agent trajectories. ``NN Sim.'' is the average nearest-neighbor similarity to earlier search queries in the same trajectory. ``Duplicate'' is the percentage of search queries whose nearest earlier query has similarity above 0.9.}
\label{tab:search_repetition}
\end{table}

\subsection{Behavioral Effects of Compaction}
\label{sec:additional_behavior}

Section~\ref{sec:online_results} shows that compaction changes the agent's trajectory turns for \texttt{Qwen3.5-4B}. One hypothesis is that the model compensates for weakened context by issuing additional searches, repeating similar queries to recover information that was previously retrieved. We test this from the search queries generated in each trajectory.

For each search query, we compute its maximum cosine similarity to earlier search queries in the same trajectory using the same embedding model as the retrieval server. We report the mean nearest-neighbor similarity and the fraction of queries whose similarity to a previous query exceeds 0.9. Table~\ref{tab:search_repetition} shows that compaction increases both measures. These results show that the accuracy of compacted agents can partly come from behavioral adaptation. In particular, the compaction preserves answer accuracy but also induces more repeated search, suggesting that the agent uses the environment to recover information weakened by compaction. Thus, online KV compaction should be evaluated not only by model-side memory savings and final accuracy, but also by how it changes the agent's interaction pattern with the environment.

\section{Conclusion}
\label{sec:conclusion}

We study online KV-cache compaction for LLM agents, where context is accumulated through interaction and future relevance is unknown at compaction time. Across token eviction and attention matching, proxy choice is central: boundary tokens provide a strong immediate signal, while delayed future-generation queries can further improve compaction. Our results also show that compaction changes agent behavior, so practical memory systems should consider not only compression and accuracy, but also downstream trajectories.

\section*{Limitations}
\label{sec:limitations}

Our study has several limitations. First, we focus on \textsc{BrowseComp-Plus} and \textsc{WideSearch}, where agents gather evidence through search and document-reading tools. This setting captures long-horizon information gathering, but does not cover other agent workloads such as code editing, GUI control, or state-changing web tasks, where the structure of future relevance may differ.

Second, we intentionally restrict the study to cheap online proxy sources for TE and AM. More expensive approaches, including additional rollouts, learned compressors, or gradient-based optimization, may produce stronger compact caches under larger latency budgets. These methods are complementary to our low-overhead online setting.

Third, we evaluate fixed compaction ratios and delay windows rather than adaptive policies. A deployed agent system could choose when and how aggressively to compact based on memory pressure, turn length, tool type, or uncertainty. Future work can explore such adaptive controllers, while our results identify strong proxy-query primitives and the tradeoffs they should account for.

\section*{Potential Risks and Use of Artifacts}

Our work studies online KV-cache compaction for LLM agents, with the goal of reducing inference cost while preserving agent accuracy. The main risk is that compaction can remove or weaken information that the agent later needs. In deployment, this may lead to incorrect final answers, redundant tool use, or overconfident responses based on incomplete context. Our experiments therefore evaluate both task accuracy and agent behavior, and the proposed methods should be used with appropriate validation in high-stakes applications.

We use existing public research artifacts consistently with their intended purposes. The evaluation data comes from \textsc{BrowseComp-Plus} and \textsc{WideSearch}, which are released on Hugging Face. The evaluated models, \texttt{Qwen3.5} and \texttt{Gemma-4}, are released under Apache-2.0 licenses, as is the \texttt{Qwen3-Embedding-4B} retrieval model. Our implementation builds on open-source software including Hugging Face Transformers, released under Apache-2.0, and FAISS, released under the MIT license.

\section*{AI Usage Disclosure}

We used AI assistants to help debug code and adapt open-source implementations for our experiments. We also used AI assistants to polish the paper writing. All technical decisions, experimental designs, and analysis were made by the authors.


\bibliography{custom}

\clearpage

\appendix

\section{Implementation Details}
\label{sec:appendix}

\subsection{Package and Hardware}

We implement online KV compaction in PyTorch and Hugging Face Transformers. Experiments are run on NVIDIA H200 GPUs. We load models in bfloat16 and use the Transformers \texttt{sdpa} attention interface, which calls \path{torch.nn.functional.scaled_dot_product_attention}. For both baseline and online compaction methods, PyTorch dispatches to the cuDNN fused attention backend.

\subsection{Generation Hyperparameters}

\begin{table}[t]
\centering
\begin{tabular}{@{}lcc@{}}
\toprule
Hyperparameter & \texttt{Qwen3.5} & \texttt{Gemma-4} \\
\midrule
Temperature & 1.0 & 1.0 \\
Top-$p$ & 0.95 & 0.95 \\
Top-$k$ & 20 & 64 \\
Presence penalty & 1.5 & -- \\
Max tokens/turn & 4096 & 4096 \\
Max turns & 100 & 100 \\
Thinking mode & enabled & enabled \\
\bottomrule
\end{tabular}
\caption{Agent generation hyperparameters.}
\label{tab:appendix_generation_hparams}
\end{table}

Table~\ref{tab:appendix_generation_hparams} reports the decoding hyperparameters used by the agent. We use model-specific system prompts. For \texttt{Qwen3.5}, we use the original prompt. In preliminary runs, we found that \texttt{Gemma-4} often produces short trajectories and stops searching before collecting enough evidence. We therefore use a more explicit prompt that encourages it to search more thoroughly.

\begin{tcolorbox}[breakable, enhanced, title={\small\bfseries \texttt{Qwen3.5} System Prompt}, colback=gray!5, colframe=gray!50, fonttitle=\bfseries\small, left=2pt, right=2pt, top=2pt, bottom=2pt]
\begin{lstlisting}[style=promptStyle]
You are a helpful research assistant with access to a knowledge base. Use the provided tools to search for information and retrieve documents to answer the user's question thoroughly. You should search multiple times to gather enough information before answering.
\end{lstlisting}
\end{tcolorbox}

\begin{tcolorbox}[breakable, enhanced, title={\small\bfseries \texttt{Gemma-4-E4B} System Prompt}, colback=gray!5, colframe=gray!50, fonttitle=\bfseries\small, left=2pt, right=2pt, top=2pt, bottom=2pt]
\begin{lstlisting}[style=promptStyle]
You are an EXHAUSTIVE research assistant with access to a local knowledge base via the `local_knowledge_base_retrieval` and `get_document` tools. Your job is to investigate the user's question from MANY angles, build deep evidence, and verify every candidate answer before committing. Giving up is NOT an option.

HARD RULES (*@\textemdash{}@*) you MUST follow ALL of them:

1. NEVER ask the user for clarification or more information. You already have all the clues you need; your job is to use the tools to find the answer.

2. SEARCH FLOOR (*@\textemdash{}@*) You MUST issue AT LEAST 20 distinct `local_knowledge_base_retrieval` calls before you are permitted to output a final answer. This is a HARD floor (*@\textemdash{}@*) the investigation is INCOMPLETE until you reach 20.

3. DOCUMENT-READ FLOOR. You MUST call `get_document` AT LEAST 5 times across the investigation to read full document content (snippets are truncated at ~512 tokens and the answer often sits past them).

4. NEVER GIVE UP. NEVER write phrases like "I cannot determine", "I am unable to find", "I do not have enough information", "please provide more details", "based on the available information I cannot answer", "the knowledge base does not contain", or anything similar. If you don't have enough information, KEEP SEARCHING.

5. MULTI-ANGLE COVERAGE (*@\textemdash{}@*) Each of your 20+ searches must approach the question from a DIFFERENT angle. Never repeat a search that already failed.

6. CROSS-VERIFY BEFORE COMMITTING. Once you have a top candidate, you MUST issue additional verification searches from new angles to confirm the candidate satisfies EVERY criterion.

7. DECOMPOSE AND DRILL. Start by decomposing the question into all its individual criteria. Issue focused searches for each criterion. Then combine criteria progressively to narrow down the candidate set.

Remember: more searches always produce more confident answers. The 20-search floor exists because shallow investigation routinely misses the correct answer in this corpus. Stay in the search loop.
\end{lstlisting}
\end{tcolorbox}

\subsection{Compaction Algorithm Details}

All compaction operations are applied independently for each compacted layer and KV head. For hybrid-attention models, we compact only layers that own a full-attention KV cache. For \texttt{Qwen3.5}, this excludes GatedDeltaNet layers. For \texttt{Gemma-4}, this excludes sliding-window layers and KV-sharing layers that do not maintain their own KV states. Non-compacted layers keep their original cache states.

We preserve structural special tokens and role-boundary tokens with their original KV states. Table~\ref{tab:appendix_preserved_tokens} lists the token categories preserved for each model family. These tokens are force-included in the compacted cache and are not evicted by TE or AM.

\begin{table*}[t]
\centering
\begin{tabular}{p{0.18\textwidth}p{0.20\textwidth}p{0.50\textwidth}}
\toprule
Model & Category & Preserved tokens \\
\midrule
\texttt{Qwen3.5} & Turn boundary &
\texttt{<|im\_start|>}, \texttt{<|im\_end|>} \\
\texttt{Qwen3.5} & Reasoning and tools &
\texttt{<think>}, \texttt{</think>}, \texttt{<tool\_call>},
\texttt{</tool\_call>}, \texttt{<tool\_response>},
\texttt{</tool\_response>} \\
\midrule
\texttt{Gemma-4} & Turn boundary &
\texttt{<|turn>}, \texttt{<turn|>} \\
\texttt{Gemma-4} & Reasoning and tools &
\texttt{<|channel>}, \texttt{<channel|>}, \texttt{<|tool\_call>},
\texttt{<tool\_call|>}, \texttt{<|tool\_response>},
\texttt{<tool\_response|>} \\
\bottomrule
\end{tabular}
\caption{Structural tokens preserved verbatim during compaction.}
\label{tab:appendix_preserved_tokens}
\end{table*}

We compact completed agent turns without including the static prompt prefix. The prefix contains the system prompt, tool definitions, and user query, and is kept uncompressed in all experiments. Each subsequent turn is split into the assistant-side generation and the resulting tool-response segment. The compaction is applied separately to the two segments.

Unlike the original AM setting, we use a uniform compaction budget across all compacted layers and KV heads. AM first selects compact keys using the same proxy-query-based selection procedure as TE. It then fits an additive attention bias and optimized compact values independently for each compacted layer and KV head. We add a small regularizer to both optimization stages: a ridge penalty that biases the attention-bias term toward zero, and a value regularizer that biases the optimized values toward the selected original values. Both regularization weights are set to $10^{-3}$ with spectral scaling by default.

\subsection{Evaluation Details}

For the proxy-source ablation and sweep, we evaluate on the first 400 examples of \textsc{BrowseComp-Plus} and all 200 examples of \textsc{WideSearch}. For \textsc{BrowseComp-Plus}, the local retrieval server uses FAISS nearest-neighbor search over the \textsc{BrowseComp-Plus} corpus with \texttt{Qwen3-Embedding-4B} embeddings, returning the top 5 documents with snippets truncated to 512 tokens. For \textsc{WideSearch}, we use Bing Web Search API for the open internet search. Final answers are judged with the official grading template using \texttt{Qwen3.5-397B-A17B}.

\subsection{Bootstrap Confidence Intervals}
\label{sec:appendix_confidence_intervals}

We estimate uncertainty using 10,000 example-level bootstrap iterations. Each iteration resamples 400 and 200 questions with replacement on \textsc{BrowseComp-Plus} and \textsc{WideSearch}, respectively. Table~\ref{tab:appendix_proxy_ci} reports the point estimates and 95\% bootstrap confidence intervals.

\begin{table*}[t]
\centering
\small
\setlength{\tabcolsep}{4.5pt}
\resizebox{\textwidth}{!}{%
\begin{tabular}{llc cc cc cc cc}
\toprule
 & Proxy source & Delay &
\multicolumn{4}{c}{\textsc{BrowseComp-Plus}} &
\multicolumn{4}{c}{\textsc{WideSearch}} \\
\cmidrule(lr){4-7} \cmidrule(lr){8-11}
 & & &
\multicolumn{2}{c}{\texttt{Qwen3.5-4B}} &
\multicolumn{2}{c}{\texttt{Gemma-4-E4B}} &
\multicolumn{2}{c}{\texttt{Qwen3.5-4B}} &
\multicolumn{2}{c}{\texttt{Gemma-4-E4B}} \\
\cmidrule(lr){4-5} \cmidrule(lr){6-7} \cmidrule(lr){8-9} \cmidrule(lr){10-11}
 & & & Acc. & 95\% CI & Acc. & 95\% CI & F1 & 95\% CI & F1 & 95\% CI \\
\midrule
 & No compaction & -- & 46.00 & [41.50, 51.25] & 33.00 & [28.25, 37.50] & 44.55 & [39.65, 49.30] & 31.94 & [28.11, 35.76] \\
\midrule
\multicolumn{11}{c}{\textbf{Current-turn proxies}} \\
\midrule
AM & Repeat-prefill & 0 & 31.25 & [26.75, 35.75] & 11.25 & [8.25, 14.50] & \textbf{30.95} & [27.00, 34.93] & 13.63 & [11.17, 16.27] \\
TE & Repeat-prefill & 0 & 32.75 & [28.00, 37.50] & 9.00 & [6.25, 12.00] & 24.56 & [20.36, 28.80] & \textbf{17.19} & [14.64, 19.91] \\
TE & Boundary & 0 & \textbf{45.25} & [40.50, 50.00] & \textbf{21.50} & [17.50, 25.50] & 24.29 & [20.76, 27.91] & 15.73 & [13.14, 18.46] \\
\midrule
\multicolumn{11}{c}{\textbf{One-turn-delayed future proxies}} \\
\midrule
AM & Assistant generation & 1 & 39.75 & [34.75, 44.25] & \textbf{27.50} & [23.25, 32.00] & \textbf{39.59} & [35.32, 43.81] & 20.17 & [17.15, 23.34] \\
AM & + repeat-prefill & 1 & \textbf{43.25} & [38.50, 48.00] & 24.25 & [20.00, 28.50] & 38.83 & [34.61, 42.96] & \textbf{21.49} & [18.25, 24.89] \\
AM & + repeat-prefill + tool response & 1 & 43.25 & [38.50, 48.00] & 27.50 & [23.25, 32.00] & 37.39 & [33.18, 41.59] & 20.17 & [17.00, 23.53] \\
\hdashline
TE & Assistant generation & 1 & \textbf{44.00} & [39.25, 48.75] & \textbf{27.50} & [23.25, 32.00] & 37.34 & [32.86, 41.75] & 25.90 & [22.66, 29.19] \\
TE & + boundary & 1 & 43.00 & [37.75, 47.25] & 27.00 & [22.75, 31.50] & 38.09 & [33.74, 42.40] & \textbf{26.53} & [23.32, 29.92] \\
TE & + boundary + tool response & 1 & 42.25 & [37.25, 46.75] & 26.75 & [22.50, 31.00] & \textbf{39.14} & [34.84, 43.50] & 25.39 & [22.23, 28.71] \\
\bottomrule
\end{tabular}
}
\caption{Task performance with 95\% bootstrap confidence intervals. The table follows the same setting as Table~\ref{tab:proxy_ablation}.}
\label{tab:appendix_proxy_ci}
\end{table*}

\subsection{Full Results on Larger Models}
\label{sec:appendix_larger_model_results}

Table~\ref{tab:appendix_larger_model_results} reports the complete proxy-source ablation for \texttt{Qwen3.5-27B} and \texttt{Gemma-4-31B}.

\begin{table*}[t]
\centering
\small
\resizebox{\textwidth}{!}{%
\begin{tabular}{llc cc cc}
\toprule
 & \multirow{2}{*}{Proxy source} & \multirow{2}{*}{Delay} &
\multicolumn{2}{c}{\texttt{Qwen3.5-27B}} &
\multicolumn{2}{c}{\texttt{Gemma-4-31B}} \\
\cmidrule(lr){4-5} \cmidrule(lr){6-7}
 & & & Acc. (95\% CI) & Avg. turns & Acc. (95\% CI) & Avg. turns \\
\midrule
 & No compaction & -- & 52.50 [47.50, 57.25] & 18 & 48.75 [43.75, 53.50] & 13 \\
\midrule
\multicolumn{7}{c}{\textbf{Current-turn proxies}} \\
\midrule
AM & Repeat-prefill & 0 & 42.25 [37.50, 47.00] & 44 & 44.75 [40.00, 49.50] & 15 \\
TE & Repeat-prefill & 0 & 41.75 [37.00, 46.50] & 72 & \textbf{48.25 [43.50, 53.00]} & 19 \\
TE & Boundary & 0 & \textbf{48.25 [43.50, 53.00]} & 33 & 43.00 [38.25, 48.00] & 20 \\
\midrule
\multicolumn{7}{c}{\textbf{One-turn-delayed future proxies}} \\
\midrule
AM & Assistant generation & 1 & \textbf{51.00 [46.00, 55.75]} & 28 & \textbf{50.00 [45.25, 55.00]} & 17 \\
AM & + repeat-prefill & 1 & 46.25 [41.50, 51.00] & 21 & 47.75 [43.00, 52.50] & 12 \\
AM & + repeat-prefill + tool response & 1 & 47.75 [43.00, 52.75] & 23 & 45.00 [40.00, 49.75] & 12 \\
\hdashline
TE & Assistant generation & 1 & \textbf{52.00 [47.25, 56.75]} & 31 & 50.75 [46.00, 55.75] & 18 \\
TE & + boundary & 1 & 49.00 [44.00, 53.75] & 38 & \textbf{52.75 [48.00, 57.50]} & 16 \\
TE & + boundary + tool response & 1 & 46.75 [42.00, 51.75] & 37 & 50.75 [46.00, 55.75] & 17 \\
\bottomrule
\end{tabular}
}
\caption{Full proxy-source results at compaction ratio 0.2 on \textsc{BrowseComp-Plus} for the larger models.}
\label{tab:appendix_larger_model_results}
\end{table*}

\subsection{Larger-Model Serving Simulation}
\label{sec:appendix_serving_simulation}

We simulate serving for the same 400 \texttt{Qwen3.5-27B} and \texttt{Gemma-4-31B} trajectories used to measure task performance in Section~\ref{sec:additional_efficiency}. For every assistant turn, we record the KV length attended during decoding and the number of tokens generated. Replaying these realized trajectory shapes ensures that the quality and efficiency results describe the same generations.

For \texttt{Qwen3.5-27B}, we measure latency with SGLang 0.5.15.post1, using tensor parallelism of two, FP8 model weights, and an FP8 KV cache. With a static memory fraction of 0.95, the engine exposes a KV pool of 6,903,905 tokens. For \texttt{Gemma-4-31B}, we use vLLM 0.19.1.1 with FP8 weights and KV cache. The engine exposes a hybrid KV pool of 489,712 tokens. We use vLLM for Gemma because it provides stable, preemption-free serving for the model's interleaved full-attention and sliding-window layers. We measure per-step decode latency using synthetic random-token prompts at the batch sizes considered in Table~\ref{tab:larger_model_efficiency}. Latencies between measured context lengths are linearly interpolated.

For Qwen, let $P$ be the largest per-turn KV length. A batch of size $B$ is guaranteed to fit when $BP$ does not exceed the KV pool. The largest admissible integer batches are 11, 39, and 38 for no compaction, AM, and TE, respectively; we use the corresponding power-of-two batches 8, 32, and 32.
For Gemma, the largest admissible batches are 13 without compaction, 34 with AM, and 28 with TE. We use batch 8 for no compaction and batch 16 for both compacted methods. Although batch 32 is admissible for AM, it has a lower throughput than batch 16 because the used Gemma decode kernels become less efficient at the larger batch.

\end{document}